\documentclass[10pt,twocolumn,letterpaper]{article}

\usepackage{authblk}

\usepackage[final]{iccv}   
\usepackage{tcolorbox}

\usepackage{bbm}
\usepackage{array} 
\usepackage{listings}
\definecolor{lightergray}{gray}{0.9} 
\usepackage{pifont} 
\newcommand{\OURS}{ScanEdit}

\definecolor{iccvblue}{rgb}{0.21,0.49,0.74}
\usepackage[pagebackref,breaklinks,colorlinks,allcolors=iccvblue]{hyperref}
\usepackage{comment}
\usepackage{stfloats}

\def\paperID{3153} 
\def\confName{ICCV}
\def\confYear{2025}

\title{\OURS: Hierarchically-Guided Functional 3D Scan Editing}

\author[1]{Mohamed El Amine Boudjoghra}
\author[2]{Ivan Laptev}
\author[1]{Angela Dai}

\affil[1]{Technical University of Munich}
\affil[2]{Mohamed Bin Zayed University of Artificial Intelligence}
\vspace{0.5em}
\affil[ ]{\url{https://aminebdj.github.io/scanedit/}}
\usepackage{mdframed}  
\usepackage{xcolor}    
\begin{document}

\twocolumn[{%
\renewcommand\twocolumn[1][]{#1}%
\maketitle

    \begin{center}
            \captionsetup{type=figure}
    \includegraphics[width=\linewidth]{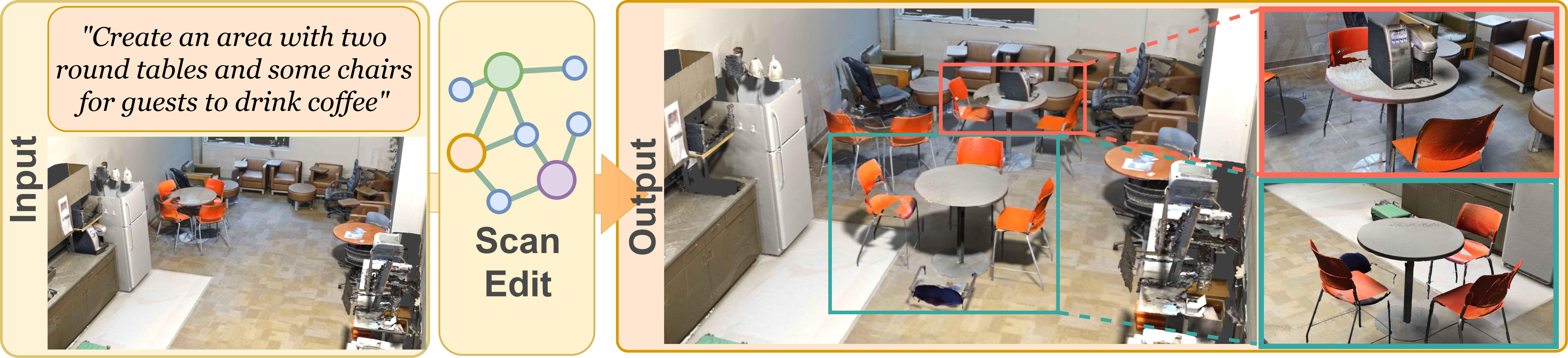}
            \captionof{figure}{
            \OURS{} enables instruction-driven editing of complex, real-world scenes by rearranging their 3D scans. Given an input 3D scan and its object-level decomposition, we formulate a hierarchically-guided, multi-stage LLM-based approach that transforms high-level user instructions into concrete, tractable local instructions for objects, which can then be globally optimized to achieve the functional instruction. In this example, ScanEdit rearranges chairs, tables and a coffee machine to create a coffee drinking area.
            }
            \label{fig:teaser}
    \end{center}
}]

\begin{abstract}
With the fast pace of 3D capture technology and resulting abundance of 3D data, effective 3D scene editing becomes essential for a variety of graphics applications.

In this work we present \OURS{}, an instruction-driven method for functional editing of complex, real-world 3D scans. 
To model large and interdependent sets of objectswe propose a hierarchically-guided approach.
Given a 3D scan decomposed into its object instances, we first construct a hierarchical scene graph representation to enable effective, tractable editing. 
We then leverage reasoning capabilities of Large Language Models (LLMs) and translate high-level  language instructions into actionable commands applied hierarchically to the scene graph.
Finally, \OURS{} integrates LLM-based guidance with explicit physical constraints and generates realistic scenes where object arrangements obey both physics and common sense. 
In our extensive experimental evaluation \OURS{} outperforms state of the art and demonstrates excellent results for a variety of real-world scenes and input instructions.

\end{abstract}

\section{Introduction}
\label{sec:introduction}

The abundance of multi-view RGB and RGB-D sensors, now available even on commodity phones, enables convenient capture and detailed reconstruction of complex real-world 3D scenes.
This opens up exciting possibilities for various applications in content creation, virtual and augmented reality, architectural design, robotics, and more. 
To fully leverage the captured 3D data, however, it is often essential to perform editing and rearrangement of the captured scenes -- for instance, re-organizing objects for content creation where iterative editing is a fundamental paradigm, or enabling a robot to visualize the goal state of the environment given the target task. 

Recent works have established the potential of leveraging powerful, pre-trained vision-language models (VLMs) for 3D scene synthesis, employing object retrieval from a synthetic 3D object database \cite{sun2024layoutvlm, feng2024layoutgpt,yang2024holodeck} to populate constructed 3D layouts to produce a 3D scene.
Such VLMs enable interpretation of a natural language prompt to construct a 3D scene layout supporting synthetic 3D object assets adhering to the prompt.
While this showcases the potential for leveraging high-level VLM-based reasoning, such generated 3D layouts populated with synthetic assets are much simpler than complex, cluttered, real-world environments (e.g., often hundreds of objects in the real-world scans we operate on, while synthetic scene synthesis approaches often generate an average number of 5 objects per scene)\cite{tang2024diffuscene,paschalidou2021atiss}). 
We thus propose the first approach to address text-based editing of complex, real-world 3D scans by leveraging a localized prompting approach which can provide reliable per-object initializations for our proposed graph optimization.
In order to handle the large  number of objects common in real 3D environments (e.g., we operate on scenes with 69-306 objects, which can easily overwhelm the context size of a VLM/LLM), we decompose the editing task into multiple stages, tackled hierarchically.
We first construct a hierarchical scene graph and identify a relevant subgraph for the input prompt.
For each object in the subgraph,  we then transform the input prompt into low-level per-object instructions grounded in the object's reference frame, which is used to generate concrete object placements, performed by iteratively traversing our hierarchical scene graph.
Finally, we use these object placements as initialization, and combine LLM-generated scene graph constraints with convex differentiable losses based on 3D object relations to achieve a physically and semantically plausible edited 3D scene.
An example of scene rearrangement generated by our method is illustrated in Figure~\ref{fig:teaser}.

Our contributions are summarized as follows:
\begin{itemize}
    \item We construct a hierarchical scene graph representation, leveraging LLM reasoning capabilities to compose graph relations, in order to enable multi-level scene analysis for functional scene editing, capable of handling complex, real-world 3D scans.
    
    \item We develop a scene graph optimization suitable for the highly diverse nature of 3D scans, employing both LLM-based object constraints in tandem with 3D-based physical constraints to avoid collisions and boundary violations, producing meaningful and plausible edited 3D scenes. 
    
\end{itemize}

\section{Related Works}
\label{sec:related_works}

\paragraph{Classical retrieval-based 3D scene modeling.}   
Creating realistic 3D scenes requires spatial understanding of object arrangements, in order to produce a 3D scene layout populated with objects from a 3D database.

One can capture such knowledge by analyzing object co-occurrence patterns, such as how often certain objects appear together in real-world environments~\cite{fisher2010context}. Another approach involves learning from human interactions, using affordance maps that show how people move and use objects within a space~\cite{fisher2015activity,fu2017adaptive,jiang2012learning,ma2016action}. To generate well-structured layouts, previous works have explored various optimization techniques, including non-linear methods that refine object placements based on spatial constraints~\cite{chang2014learning,qi2018human,xu2013sketch2scene,yeh2012synthesizing,yu2011make,fisher2012example}. Although these methods can generate plausible layouts, they do not support complex reasoning required for interpreting high-level natural language prompts to enable rearrangement of 3D scans.

\bigskip 
\noindent
{\bfseries Learning for retrieval-based 3D scene synthesis.}

Advances in deep learning encouraged adoption of various techniques for 3D scene synthesis, using feed-forward networks \cite{zhang2020deep,wu2022targf}, VAEs \cite{purkait2020sg,yang2021scene}, GANs \cite{yang2021indoor}, and autoregressive models \cite{xiao2021tackling}. A popular approach is to employ autoregressive models \cite{li2019grains,ritchie2019fast,wang2019planit,wang2018deep,wang2021sceneformer} to predict object arrangements in a sequential fashion, and explicitly model relations among objects at the cost of spatial complexity.
 
 Diffusion-based methods \cite{sohl2015deep,song2019generative,song2020improved,song2020denoising,ho2020denoising} overcome some of these limitations by modeling object distributions holistically. Recent methods such as LEGO-Net~\cite{wei2023lego}, CommonScene~\cite{zhai2024commonscenes}, and DiffuScene~\cite{tang2024diffuscene} further improve scene plausibility and coherence. These methods excel at unconditional scene generation, but focus on relatively simple synthetic scenes with 10-20 objects maximum, while we develop a hierarchical approach to tractably address complex, real-world scenes with hundreds of objects.

\smallskip 
\noindent
{\bfseries 3D shape and scene editing.}
3D editing has been addressed by a number of works.

Most methods focus on 3D shape editing~\cite{instructnerf2nerf, tailor3d, mvedit,voxe, shapeditor, interactive3d, paintbrush, erkocc2024preditor3d}, while several more recent approaches address a more complex task of editing larger-scale 3D scenes~\cite{bokhovkin2024scenefactor, Haque_2023_ICCV, ocal2024sceneteller, Sun_2024_WACV, Wang_2024_CVPR}. 

Such method focus on flexibility in localized editing, which in scenes can often alter object instance identities. On the contrary, we focus on editing the original scene mesh representation by re-arranging objects to adhere to high-level, functional text prompts.

\smallskip 
\noindent
{\bfseries LLMs/VLMs for retrieval-based 3D scene synthesis.}

Another recent direction of work explores 3D scene generation by synthesizing intermediate representations, such as scene graphs or layouts, paired with an asset repository \citep{rahamim2024lay, feng2024layoutgpt, fu2025anyhome, yang2024holodeck, lin2024instructscene, zhou2024gala3d}. The rise of Large Multimodal Models (LMMs) has opened up new possibilities for open-vocabulary 3D scene synthesis, allowing for the creation of scenes without being constrained by predefined labels or categories \cite{ccelen2024design,aguina2024open}. For instance, LayoutGPT \cite{feng2024layoutgpt} uses language models to directly produce 3D layouts for indoor scenes. Similarly, Holodeck \citep{yang2024holodeck} employs large language models (LLMs) to generate spatial scene graphs, which are then used to optimize object placements. Another approach, LayoutVLM \cite{sun2024layoutvlm}, adopts a differentiable optimization method to create scene layout representations from image and text inputs using Vision-Language Models (VLMs). A recent work, Fireplace \cite{huang2025fireplace}, addresses object placement in 3D scenes by leveraging large multimodal models (LMMs) and solving geometric constraints to achieve a logical placement of new objects. 

Unlike previous methods which follow an object retrieval approach to scene synthesis, Gala3D \cite{zhou2024gala3d} uses an LLM to generate a possible arrangement for relevant objects in the input text instruction, followed by Gaussian splatting to generate the target objects and optimization achieve a plausible layout of objects. 
In contrast, we aim to perform functional editing of real-world 3D scans, which are often complex, with far more objects (our scenes contain 69-306 objects), requiring our hierarchical approach to characterize tractable editing while employing LLMs and VLMs. 
\begin{figure*}[ht]
    \centering
    \includegraphics[width=\linewidth]{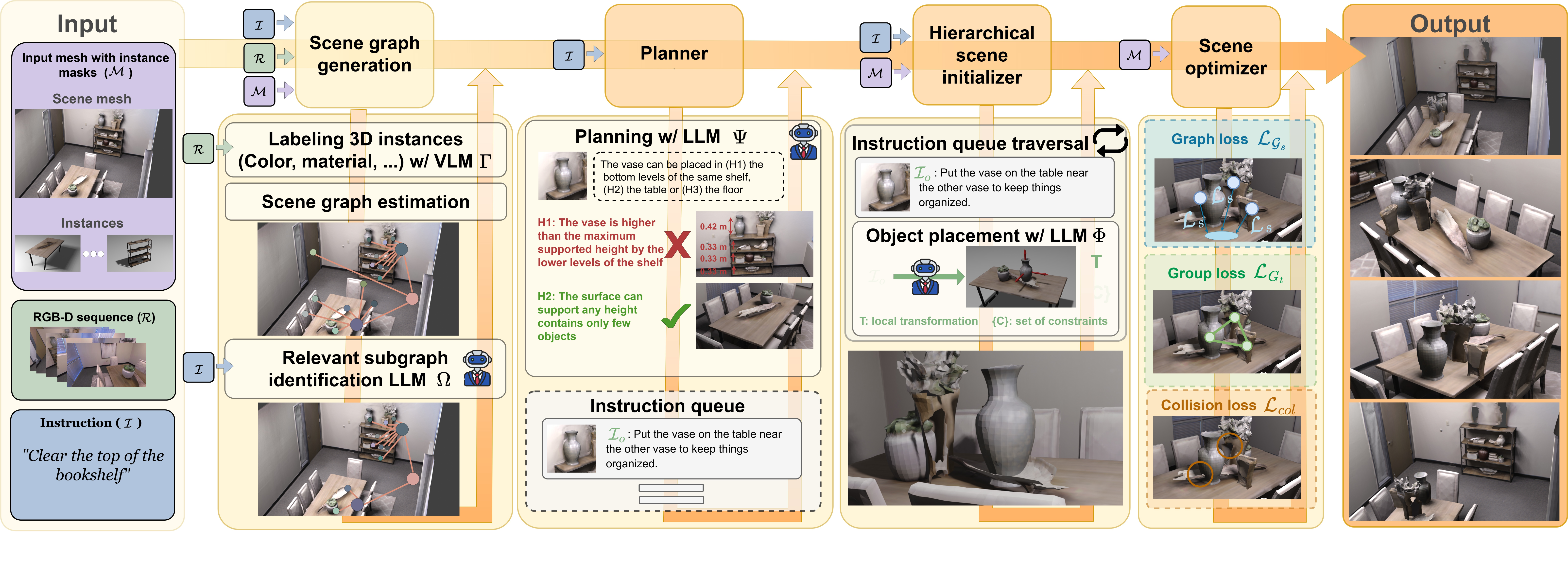}
    \vspace{-0.6cm}
    \caption{Overview of \OURS. 
    Given an input instruction $\mathcal{I}$ for a scene mesh $\mathcal{M}$ that has an instance decomposition and is reconstructed from RGB-D sequence $\mathcal{R}$, we output an edited scene according to the instruction $\mathcal{I}$. 
    We first construct a hierarchical scene graph $\mathcal{G}$ using 3D and VLM reasoning to annotate graph node and edge attributes. Since this graph may be very large in size, we then identify the relevant subgraph $\mathcal{G}_s$ for $\mathcal{I}$.
    Our planner then breaks down the high-level instruction $\mathcal{I}$ into low-level object instructions, validates them, and creates an instruction queue.
    We traverse the instruction queue hierarchically in order to place objects as initialization for the new output scene, followed by a scene optimization over both LLM-generated scene constraints as well as physical 3D collision constraints, to produce the output edited scene.
    In this example, the two vases on the top of the bookshelf are moved to the table in the 3D scan.
    Note that since real-world 3D scans are partial, some holes can be visible (e.g., in the output wall) after re-arranging objects.
    }
    \label{fig:method}
\end{figure*}
\section{Method}

Given an input natural language text instruction $\mathcal{I}$ to edit a 3D scan $\mathcal{M}$ reconstructed from an RGB-D sequence $\mathcal{R}$ and decomposed into $N$ semantic instances $\{o_1,...,o_N\}$, we aim to predict an edited, rearranged scene $\mathcal{M}'$  as output based on $\mathcal{I}$.
The edited output scene $\mathcal{M}'$  is composed by re-arranging the object instances $\{o_i\}$ with new locations and orientations defined as transformations $T_{i}\in\mathbb{R}^{4\times 4}$. The $T_{i}$ must lead to a physically plausible (without colliding or floating objects) scene configuration, while adhering to the instruction $\mathcal{I}$. 

To achieve the above goal, we first construct a scene graph $\mathcal{G} = (N, E)$ representing $\mathcal{M}$, where we deploy a VLM $\Gamma$ to annotate attribute features for the graph nodes $N$ representing each object, as well as annotate the object-object edge hierarchical relations represented by $E$ (Sec.~\ref{subsec:scenegraph}).
As the full scene graph $\mathcal{G}$ is typically large and contains hundreds of objects, we then identify the subgraph $\mathcal{G}_s$ relevant to $\mathcal{I}$ using an LLM agent (Sec.~\ref{subsec:subgraph}).
For each object in $\mathcal{G}_s$, we then decompose $\mathcal{I}$ into localized, object-specific instructions that should be performed to achieve $\mathcal{I}$ (Sec.~\ref{subsec:planning}). 
These instructions are then interpreted through our hierarchical object placement for an initial output scene (Sec.~\ref{subsec:placement}). 
Finally, we optimize the edited scene $\mathcal{M}'$ 
by combining LLM-generated functional scene constraints with physical 3D scene constraints.

An overview of our method is presented in Figure.~\ref{fig:method}.

\subsection{Hierarchical Scene Graph Construction}
\label{subsec:scenegraph}

Our hierarchical scene graph $\mathcal{G} = (N, E)$ provides a compact and holistic representation for scene $\mathcal{M}$  to facilitate scene editing. 
Each node in $\mathcal{G}$ represents an object in $\mathcal{M}$, and the scene hierarchy is established based on several intra-object edge relationships: 
`on top of', `facing', and `against wall.'
Initially, nodes are simply given by the object instance masks of the geometry, and we further estimate node attributes and edges from the input scene information.

\noindent\textbf{Node-specific attributes}: Each object node \( o_i \in N \) is characterized by the following:
{Class} $c_i$; Color $\chi_i$; {Material} $m_i$; {Short Description} $d_i$; {Front Normal} $\Vec{F}_i$; {Point Cloud} $P_i$ and {Support Surfaces} 
$S_i$.

\noindent \textbf{Node-to-node relations}:
We encode these as directed edges \(e^{o_i\rightarrow o_j} = o_i \rightarrow o_j \) in the scene graph, with possible edge types as `on top of', `facing', `against wall. $e_{On-top-of}^{o_i\rightarrow o_j}$ represents the relation where the object $o_i$ is on top of one the surfaces of object $o_j$, $e_{Against-wall}^{o_i\rightarrow o_j}$ represents the relation where the object $o_i$ is against the wall $o_j$, and $e_{Facing}^{o_i\rightarrow o_j}$ represents the relation where the object $o_i$ is facing $o_j$.

\paragraph{Estimating node-specific attributes.}
Node attributes are annotated using a VLM agent $\Gamma$. 
To provide $\Gamma$ with sufficient information, we use the RGB-D sequence $\mathcal{R}$ associated with the 3D scan $\mathcal{M}$ and rasterize 2D masks of  3D objects. Resulting image crops are used as input to VLM.

\noindent {\textit{Color $c_i$, Material $m_i$, Short description $d_i$:}}  These are all text attributes estimated by $\Gamma$. 
They describe the most dominant object color and material, as well as a short description of fine-grained details. %
\noindent {\textit{Sampled points $P_i$:}}
We describe the object geometry with sampled points on its surface at $\approx1$cm resolution.

\noindent {\textit{Support surfaces $S_i$:} } are the regions of $o_i$ that could potentially support another object. 

A support surface $S_i$ is defined by its set of contour points $S_{(i, c)}$ that define the boundaries of the surface, and the set of normals per contour point $S_{(i, \Vec{n})}$ that point to the outside of the surface.

\noindent {\textit{Estimating surface contour points $S_{(i, c)}$ and normals $S_{(i, \Vec{n})}$:}  }
For a given surface level, we first identify contour points as those farthest from the barycenter. To estimate normals, we fit a spline interpolation to the contour, providing a smooth function from which we can sample points for improved normal estimation. These contour points and normals are then used to compute boundary loss in our method.

\noindent {\textit{Front normal $\Vec{F}_i$:} }
is used to understand the orientation of the object in the 3D space, and inform relations such as `facing' or `against wall.'  
As some objects have a clear front-facing direction (e.g., an armchair), we classify if they should have such a semantic front-facing direction with an LLM and use its oriented bounding box to estimate the facing direction. 
For other objects, we populate $\Vec{F}_i$ with the  axis of symmetry that best aligns with the average of the object's vertex normals. 

\paragraph{Estimating edges:} We introduce an edge into the scene graph between nodes $(o_i, o_j)$, based on 3D spatial reasoning in the scene to estimate the relations `on top of', `facing', and `against wall'. We refer to the supplemental for further details.

\subsection{Relevant Sub-graph Identification}
\label{subsec:subgraph}

Real 3D scans typically contain hundreds of objects, leading to thousands of relationships between them. This large amount of scene context can easily overwhelm the limited context window of a LLM agent. 
However, we observe that much of a scene graph may be irrelevant for a given instruction $\mathcal{I}$.
To simplify planning and reduce hallucinations, we first employ LLM agent $\Omega$ to filter the input graph $\mathcal{G}$, identifying only the relevant objects and relationships, which results in a subgraph $\mathcal{G}_s\subseteq\mathcal{G}$ as the input for the planner. Relevant nodes include objects that need to be moved (including their children in the scene graph) and their potential target locations, while relevant relationships depend on spatial prompts in the instruction. 

\subsection{Localized Planning}
\label{subsec:planning}

Given the relevant subgraph $\mathcal{G}_s$ for the instruction $\mathcal{I}$, we employ a planner LLM $\Psi$ to generate a new edited subgraph $\mathcal{G}'_s$ that should represent the edited scene according to $\mathcal{I}$. This is done by reconnecting edges of $\mathcal{G}_s$ to form the edited scene graph, as well as generating corresponding node-level instructions in the local reference frame of its parent. 

Our LLM planner $\Psi$ is invoked for each object node in $\mathcal{G}_s$, and prompted to generate a set of hypotheses as to how it should be transformed to adhere to $\mathcal{I}$, and select the best out of the generated hypotheses.  
This multiple hypothesis generation and selection helps to maintain robustness against hallucinations and incorrect hypothesis generation.
To promote the best outcome, we prompt $\Psi$ to generate multiple hypotheses for object-level text instructions $\{I^{o_i}_k\}$ and select the most suitable one $I^{o_i}$. The LLM $\Psi$ receives the sub-graph $\mathcal{G}_s$ to be edited, with each node described with its geometric information: \textit{dimensions}, \textit{maximum height for support surfaces} and \textit{objects on top of each surface}. We prompt the LLM $\Psi$ to take into account these details generating each object's hypotheses and select the best one based on geometric plausibility.  
The object-level instruction hypotheses $\{I^{o_i}_k\}$ are generated in the local reference frame of the parent of $o_i$ in order to localize focus during the instruction generation.

The output of the planner will consist of an edited subgraph $\mathcal{G}_s'$, where each node has a corresponding instruction $I^{o_i}$ in natural language.

\subsection{Hierarchical object placement}
\label{subsec:placement}

Our hierarchical placement agent, LLM $\Phi$, is called multiple times for each node in the edited sub-graph  $\mathcal{G}'_s$. Given an object $o_j$ in $\mathcal{G}'_s$, it determines the placement of its children relative to it. For each child, the agent generates an $(x,y,z)$ position along with an orientation $\theta$, specifying the desired pose for all children in $\textbf{Children}(o_j)$. We denote the relative position and orientation with the matrix transform $T$ (also shown in Figure~\ref{fig:method}). Additionally, it considers possible constraints (denoted $\{C\}$), selected from `against wall,' `on top of,' and `facing,' which are optimized during the optimization step.

\subsection{Scene Subgraph Optimization}
\label{subsec:optimization}

In this section, we introduce a set of convex differentiable loss functions designed to resolve graph edge constraints generated by the Placement LLM $\Phi$. The surface loss ensures that an object stays within the boundaries of a designated support surface, while the against-wall loss ensures that the object is positioned correctly against a target wall (if applicable). We also include a geometric collision loss, which helps resolve any collisions between the object and nearby objects. In general, we introduce a geometric group loss that preserves the structure of the groups during optimization. In our method, the `facing' constraint is resolved in the hierarchical placement step where the child is placed to face the parent if 'facing' is required, while the other two edges types `on top of' and `against wall' are resolved in the optimization phase. 

\paragraph{\textbf{"On-top-of" loss.}}

Since LLMs are known to  lack spatial awareness during reasoning, the Object Placement LLM $\Phi$ occasionally places objects in out-of-surface boundary locations to avoid collision with objects on the support surface. To ensure that objects are in their designated support surfaces, we introduce a loss to encourage an object to stay in the target support surface defined by constraint `on top of' generated by the LLM.

Given an object $o_i$ and a target surface $S_j$, we define the surface loss as the signed distance between the sampled points $P_i$ of object $o_i$ to $S_j$. The signed distance for a point $p$ to $S_j$ can be computed as $d(p) = (p-p_s)\cdot \Vec{n_s}$ where $p_s$ and $\Vec{n}_s$ are the closest point in the contour points $S_{j, c}$ and the normal $S_{j, \Vec{n}}$ of that contour point in $S_j$ respectively. Finally, the loss is the average across all points:
$$
\mathcal{L}_\textrm{On-top-of}(P_i, S_j) = \frac{1}{N}\sum_{p \in P_i} \max(0, d(p)).
$$

\paragraph{\textbf{"Against-wall" loss.}}
Since some objects are naturally placed against walls, if the target location for an object is specified to be against a wall, we employ a loss to encourage the object to remain against its target wall during optimization.

Given a wall defined by its world-to-wall transformation \( T_\textrm{wall} \) and oriented bounding box (OBB) extents \( Dx_\textrm{wall}, Dy_\textrm{wall} \), and an object $o_i$ to be placed against the wall defined by its OBB extents \( Dx_o, Dy_o \) and center $c_{o_i} = [c_{o_i, x}, c_{o_i, y}]$ where $c_{o_i, x}, c_{o_i, y}$ qre the \textit{x}, \textit{y} coordinates of the object $o_i$ in the world coordinate system respectively. We initialize the object $o_i$ with the same orientation as the wall, aligning their coordinate frames. The object's center is then adjusted to be positioned at a fixed distance from the wall, ensuring that the gap between their centers along the \( x \)-axis in the wall's coordinate frame is exactly \( \frac{Dx_\textrm{wall} + Dx_{o}}{2} \). This guarantees that the object remains in contact with the wall without penetrating it. 

After transforming both the wall and the object into the wall coordinate frame, they are aligned to face the positive \( x \)-direction. The object's position is then optimized to remain within \( \pm \frac{Dy_\textrm{wall}}{2} \) along the \( y \)-axis in this frame.  
The object's center in this coordinate frame is computed as:  
\[
c_{o_i}^\textrm{wall} = c_{o_i} \cdot T_\textrm{wall}^{-T}
\]
with OBB extents \( Dx_o, Dy_o \). The \textbf{against-wall loss} is then defined as:  
\[
\mathcal{L}_\textrm{AgainstWall} =
||c_{o_i, x}^\textrm{wall} - \frac{Dx_\textrm{wall} + Dx_o}{2}|| + ||c_{o_i, y}^\textrm{wall}||\cdot \text{OPT}
\]
where
\[
\text{OPT} = \mathbbm{1} \left( |c_{o_i, y}^\textrm{wall}| > \frac{Dy_\textrm{wall}}{2} \right).
\]
This formulation allows the object to slide freely along the \( y \)-axis within the boundaries of the wall.
\paragraph{\textbf{Collision loss.}}

LLM-generated placements often lead to collisions between objects. To resolve collisions during optimization, we propose the loss to discourage collisions.

Given the set of nodes $N_t \in N$ to be optimized for, where $N$ are the nodes of the graph $\mathcal{G}$ and $N_t$ are the nodes of $\mathcal{G}_s$, for all objects $o_i \in N_t$, we optimize a collision loss $\mathcal{L}_{\textrm{col}, i}$ that pushes the object away from other object geometry in the scene.
Specifically, the center of $o_i$ should move away from the centers of other objects in $\mathcal{G}_s$, and any points belonging to other objects in $\mathcal{G}$ that lie in the bounding box of $o_i$ should be pushed away from $o_i$'s points $P_i$.
We denote the set of points belonging to other objects' surfaces that lie in the bounding box of  $o_i$ as $P_{o_i}$. 
$$
\begin{aligned}
\mathcal{L}_{\textrm{col}, i}^{push} &= (\frac{1}{n} \cdot \sum_{d_i \in d(P_i , P_{o_i})}\max(0, \Delta - d_i) \\
&\quad + \frac{1}{n_t-1} \cdot \sum_{j \in N_t, j\neq i} \max(0,\Delta - ||c_i - c_j||)),
\end{aligned}
$$
where $c_i$ is the center of the $i^{th}$ object, $d: (\mathbb{R}^{n_1\times 3}, \mathbb{R}^{n_2\times 3})\rightarrow \mathbb{R}^{n_1\times n_2}$ is a distance function evaluating all pairwise distances between two sets of point clouds, $n = n_1\times n_2$, and $n_t$ is the total number of objects to be optimized for.

Finally, we multiply the push loss $\mathcal{L}_{\textrm{col}, i}^\textrm{push}$ by a stop condition $\text{STOP}_{col}$ which sets the loss to 0 if the object is not colliding with any other objects anymore. We measure collision by the minimum distance between the $P_i$ and the $P_{o_i}$,  where $r$ is the resolution of the points $P_i$. which makes the final collision loss
$$
\mathcal{L}_{\textrm{col}, i} = \mathcal{L}_{\textrm{col}, i}^\textrm{push}\cdot\text{STOP}_\textrm{col},
$$
where $$
\text{STOP}_\textrm{col} = \mathbbm{1}(\min(d(P_i, P_{o_i})) < 4\cdot r).
$$
The collision loss across all objects $N_t$ is
$$
\mathcal{L}_{\textrm{col}} = \sum_{o_i \in N_t} \mathcal{L}_{\textrm{col}, i}.
$$
\paragraph{\textbf{Group loss.}}

We define object groups by the sets of objects that share the common support surface and are assigned to the same parent in the scene graph. Generally, LLMs are capable of producing good arrangement of objects (e.g., chairs in a circle or vases in a line) but some of these objects can collide with other objects or be placed out of boundary. In order to preserve the structure of these groups during optimization, we introduce a loss that preserves the relative vectors between object centers similar to the initialization, and keeps group members in a distance similar to the initial distance to their group center.

Given a set of objects $\{o_0, o_1, o_2\}$ which belong to the same group where $g_c$ is the parent node in the graph, and their updated center at a certain step $t$, $G_t = [c_{o_0, t}, c_{o_1, t}, c_{o_2, t}]$. The group loss is defined as:

$$
\mathcal{L}_{G_\textbf{t}} = \sum_{(o_{i, \textbf{t}}, o_{i, \textbf{0}}) \in (G_\textbf{t}, G_\textbf{0})}\sum_{(o_{j, \textbf{t}}, o_{j, \textbf{0}}) \in (G_\textbf{t}, G_\textbf{0})} ||v_{i\rightarrow j, \textbf{0}}-v_{i\rightarrow j, \textbf{t}}||,
$$
where $v_{i\rightarrow j, \textbf{t}}$ is the directional vector from object $o_{i, \textbf{t}}$ to $o_{j, \textbf{t}}$, and is defined as $v_{i\rightarrow j, \textbf{t}} = c_{o_j, \textbf{t}}-c_{o_i, \textbf{t}}$.

\paragraph{Final optimization objective.}
For a set of constraints $\{C\}$  generated by the Placement LLM, the final objective is a weighted sum of the losses of all constraints $\{C\}$ that is described by the graph loss $\mathcal{L}_{\mathcal{G}}$ , the group loss $\mathcal{L}_{G_t}$, and the collision loss $\mathcal{L}_\textrm{col}$.

$$
\mathcal{L} =
\mathcal{L}_{\mathcal{G}_s}+\alpha \mathcal{L}_{col}+\gamma \mathcal{L}_{G_t}.
$$

\paragraph{Gradient Update.} 

 For a given object \( o_i \) placed on a support surface, we initialize its height to match that of the surface. This reduces the optimization to only the \( x, y \) translation and rotation around the \( z \)-axis. We optimize the objective function $\mathcal{L}$ using stochastic gradient descent (SGD) with a cosine annealing learning rate scheduler. For an object $o_i$ with dimensions $D x_i$ and $D y_i$ along the $x$- and $y$-axes, respectively, we optimize a transformation $T_i$ with three degrees of freedom: translation along $x$ and $y$, and a rotation by angle $\theta$.

Each object $o_i$ is assigned a scheduler with a maximum step size in the $x$-direction given by $lr_{i, x, \max} = 0.25 \cdot D x_i$ and in the $y$-direction given by $lr_{i, y, \max} = 0.25 \cdot D y_i$. This scheduler ensures that the step size is adapted to each object's dimensions, allowing for a more flexible and controlled optimization process.

\label{sec:results}
\section{Experimental setup}
Our experimental setup includes eight scenes, with four from the ScanNet++ validation set and four from Replica.
We use on average 12 instructions per scene, totaling 93 evaluation samples. 

\paragraph{Evaluation Metrics.}
We assess performance using both perceptual and geometric metrics. The perceptual evaluation is based on the judgment of 25 human subjects. It comprises a binary component, where users choose the best result of two given methods, and a unary component, where results of each method are rated on a five-point scale. 
For the geometric evaluation, we adopt PIoU from DiffuScene~\cite{tang2024diffuscene} to measure collisions. Since bounding box IoU only provides a coarse collision estimate, we also introduce the ColVol measure that approximates the cumulative volume of colliding object parts in a scene.

We also evaluate the percentage of objects that are not floating (NoFloat), as well as the percentage of objects within the bounds of the scene (InBound).
For further details on the evaluation metrics we refer to the supplemental.

\paragraph{Baselines.}
We compare our method with state-of-the-art methods LayoutGPT~\cite{feng2024layoutgpt} and LayoutVLM \cite{sun2024layoutvlm}. LayoutGPT relies on in-context samples retrieved to be similar to the input instruction, to generate a layout for retieved 3D assests. LayoutVLM uses a VLM for understanding the current state of the layout of 3D objects along with  rendered images of the 3D scene to generate an arrangement of objects. This arrangement is obtained through VLM 3D intialization followed by differentiable optimization for a set of constraints also generated by the VLM. 
\section{Results}

 \begin{table}[b]
    \centering
    \caption{Our method outperforms both LayoutGPT~\cite{feng2024layoutgpt} and LayoutVLM~\cite{sun2024layoutvlm} in all geometric plausibility measures, due to our hierarchical approach and joint 3D-based scene optimization.}

    \label{tab:main_results}
    \resizebox{\linewidth}{!}{
        \begin{tabular}{lcccc}
            \toprule
            Method & NoFloat (\%) $\uparrow$ & InBound (\%) $\uparrow$ & ColVol ($m^3$)  $\downarrow$& PIoU  $\downarrow$  \\
            \midrule
            LayoutGPT~\cite{feng2024layoutgpt} & 53.65&63.91 &1.3440& 0.478    \\
            LayoutVLM \cite{sun2024layoutvlm} & 67.55&85.74 & 1.4441& 0.491   \\
            \textbf{ScanEdit (Ours)} & \textbf{88.41} & \textbf{99.77}& \textbf{1.3381}& \textbf{0.472} \\
            \bottomrule
        \end{tabular}
    }
\end{table}

\begin{figure*}[tp]
    \centering
    \begin{tabular}{>{\centering\arraybackslash}m{0.2\textwidth} 
                    >{\centering\arraybackslash}m{0.2\textwidth} 
                    >{\centering\arraybackslash}m{0.2\textwidth} 
                    >{\centering\arraybackslash}m{0.2\textwidth}}
        \textbf{Input} & \textbf{LayoutVLM} & \textbf{LayoutGPT} & \textbf{Ours} \\
    \end{tabular}
    \\
    \includegraphics[width=0.9\textwidth]{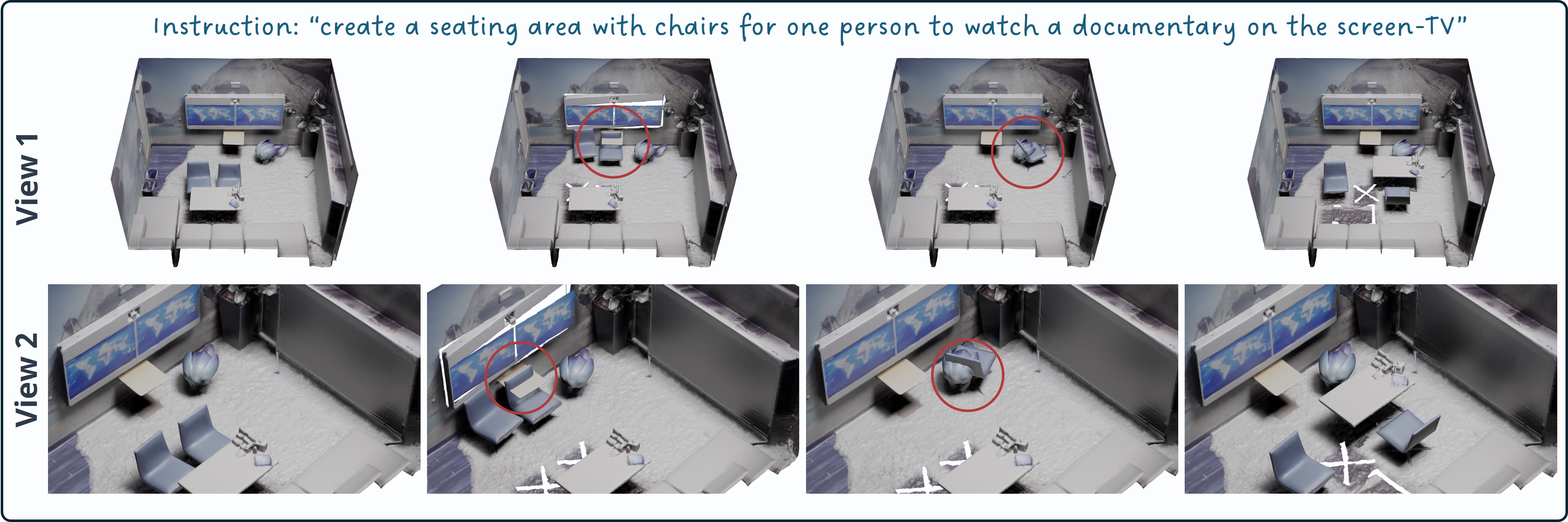} \\
    \includegraphics[width=0.9\textwidth]{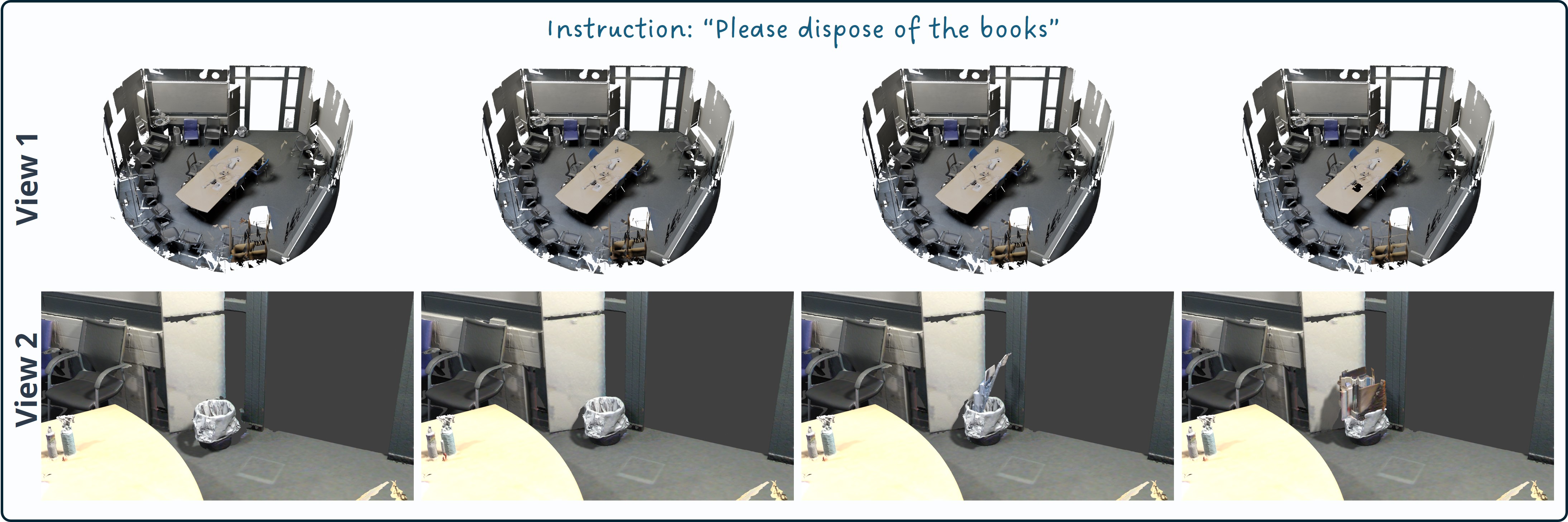} \\
    \includegraphics[width=0.9\textwidth]{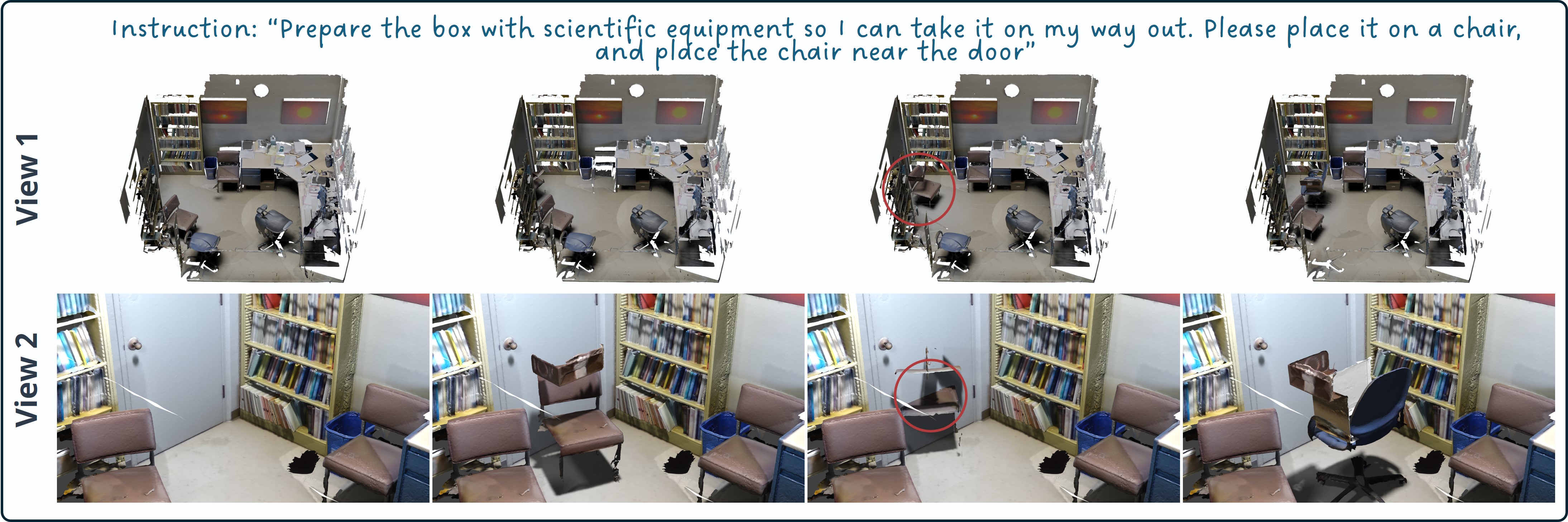} \\
    \includegraphics[width=0.9\textwidth]{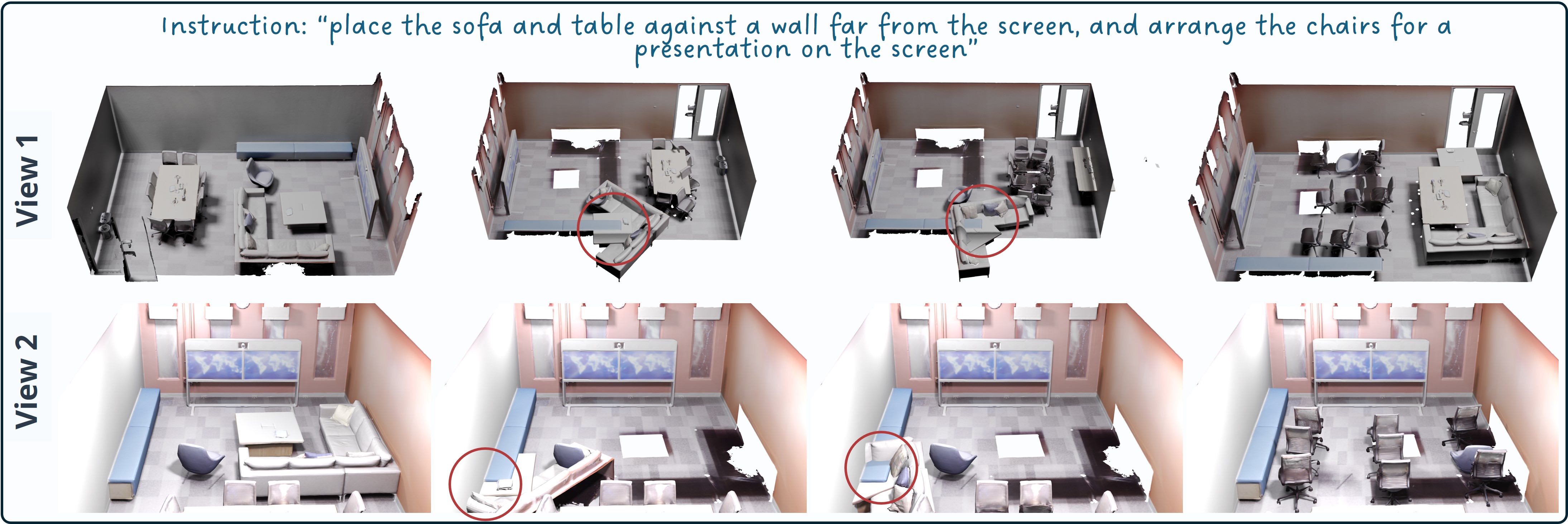}
    \caption{Qualitative comparison with baselines LayoutGPT~\cite{feng2024layoutgpt} and LayoutVLM~\cite{sun2024layoutvlm}. Red circles  denote strong geometric errors (large collisions, out-of-boundary).  Our method shows strong performance in adhering to the instruction while achieving physical plausibility.}
    \label{fig:all_figures}
\end{figure*}

\begin{figure*}[tp]
    \centering
    \includegraphics[width=\linewidth]{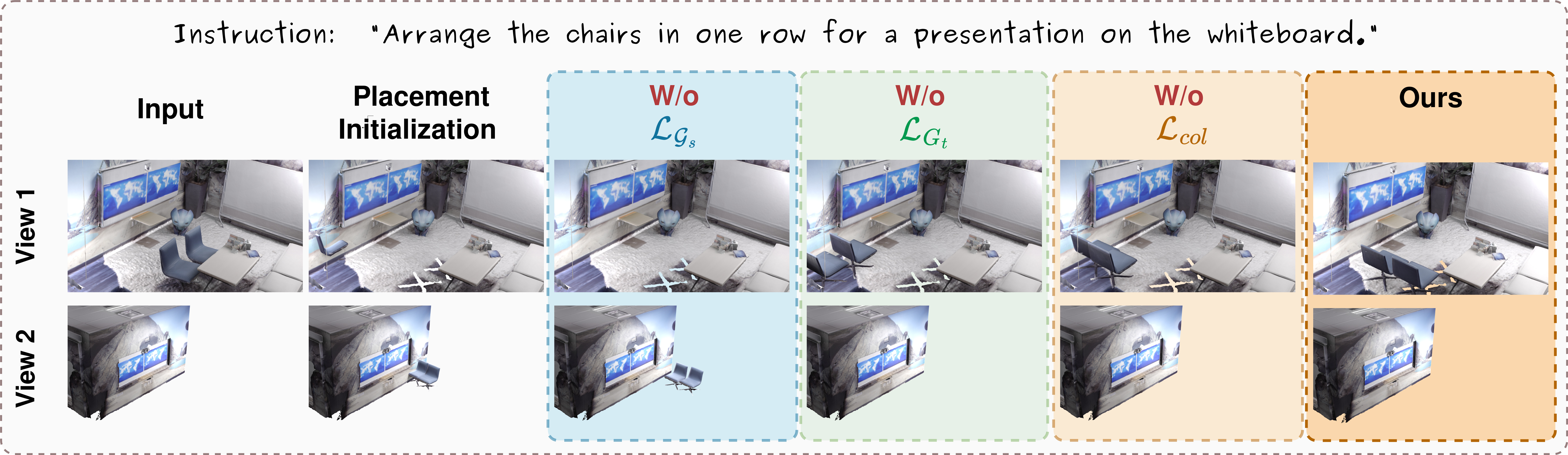}
    \caption{Ablation visualization over loss components. Our final loss with all components produces physically plausible results that avoid collisions and out-of-boundary objects. }
    \label{fig:ablation}
\end{figure*}
\begin{figure}[bp]
    \centering
   \includegraphics[width=\linewidth]{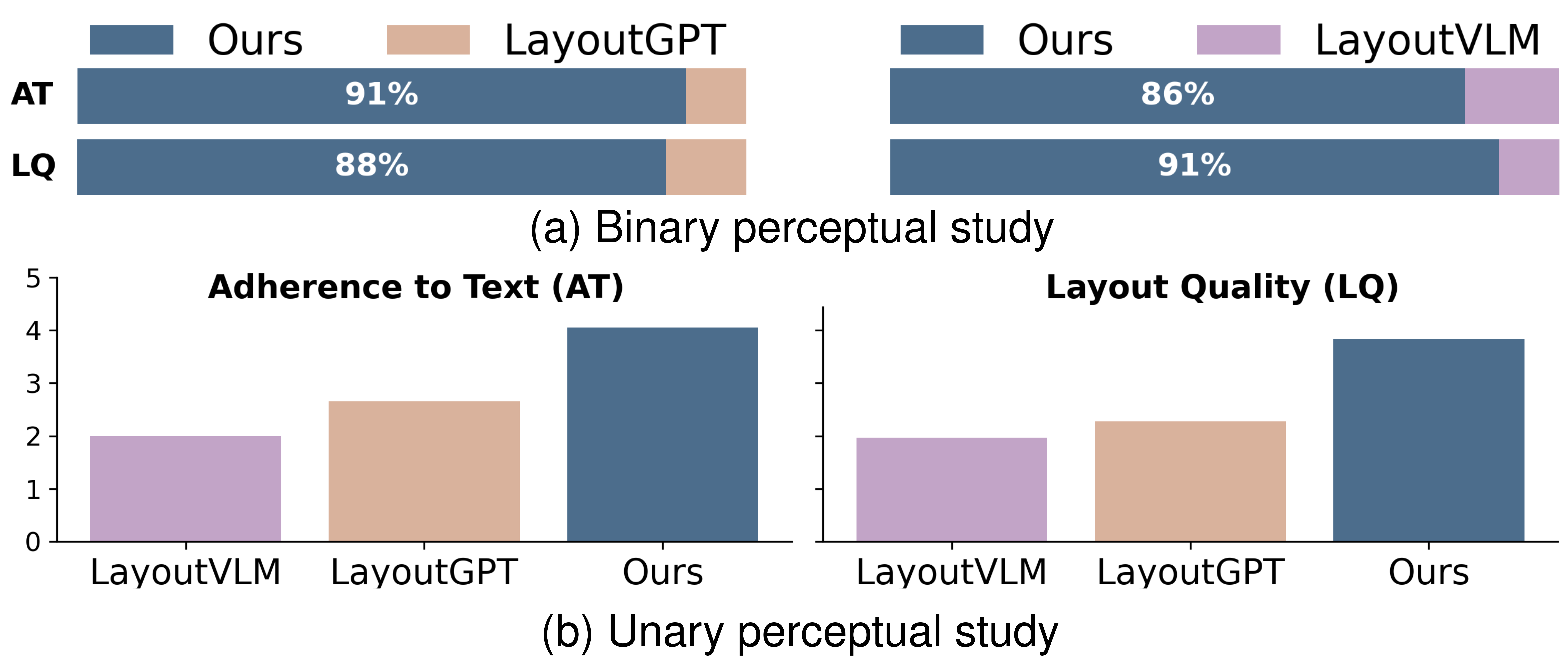}

    \caption{Our perceptual study shows that users strongly prefer our method compared to baselines LayoutGPT~\cite{feng2024layoutgpt} and LayoutVLM~\cite{sun2024layoutvlm}, in both of adherence to text instruction (AT) and layout quality of the edited scene (QL).}
    \label{fig:perceptual_study}
\end{figure}

\subsection{Comparison to state of the art}
Table~\ref{tab:main_results} shows a comparison to state-of-the-art methods LayoutGPT~\cite{feng2024layoutgpt} and LayoutVLM~\cite{sun2024layoutvlm}, evaluating geometric plausibility of edited scenes from ScanNet++~\cite{yeshwanthliu2023scannetpp} and Replica~\cite{straub2019replica}.
Our hierarchically-guided approach outperforms both baselines across all metrics.
In particular, LayoutGPT can suffer from lack of spatial understanding inherent to its reliance on LLMs, and LayoutVLM can struggle to handle complex scenes with many objects.

Figure~\ref{fig:all_figures} shows a qualitative comparison. Both LayoutGPT and LayoutVLM have difficulty handling the complex nature of real scan 3D scene environments, resulting in more collisions and misplaced objects. In contrast, our hierarchical decomposition enables more semantically and geometrically plausible edited scene outputs.
This is further confirmed by our perceptual study, shown in Figure~\ref{fig:perceptual_study}, where our method is strongly preferred by participants, who also rate our edited scene results as having notably better layout quality and adherence to text inputs.

\subsection{Ablation studies}
\paragraph{What is the impact of using a hierarchical graph (H-Graph)?} 
By comparing Row ID 2 to Row ID 1 in Table~\ref{tab:ablation}, we examine the effects of using a hierarchical graph structure as input to the LLM rather than just node-level properties. Prompting the LLM without encoding edge relations between objects is represented by row ID 1. According to our findings, performance on the geometric metrics slightly declines in this setting. We explain this by the fact that when edge information is not included, the prompt length is shortened. Given their small context window, LLMs typically reason better when given brief but instructive prompts.

\paragraph{What is the impact of the planner (Plan)?}  
We can see that merely introducing a planner is insufficient by comparing Row ID 3 to Row ID 2 in Table~\ref{tab:ablation}. Even with a rationally sound plan, the placement LLM still has trouble positioning several objects at once, which produces less-than-ideal results.

\paragraph{What is the impact of hierarchical placement (H-Place) when used with Plan and H-Graph?} 
We also consider a simplified variant of our approach that relies on an LLM to directly generate object transforms for the relevant objects to be moved (Row ID 5 compared to Row ID 1 in Table~\ref{tab:ablation}). 

This yields fairly low geometric performance, primarily due to the LLM's limited 3D spatial understanding. This leads to more objects being placed outside the room boundaries, as reflected in the low InBound score, as well as increased collision cases, indicated by higher ColVol and PIoU values. Additionally, many objects are left floating, as seen in the low NoFloat score. 
In contrast, our proposed hierarchical planning and placement, followed by scene optimization, significantly improves output scene plausibility, see also Figure~\ref{fig:ablation}.

\paragraph{What is the impact of the joint scene optimization?} 
In Table~\ref{tab:ablation} we evaluate our approach in comparison to a variant without the final scene optimization (Row ID 6 compared to Row ID 5 in Table~\ref{tab:ablation}).
Our scene optimization shows a notable improvement, as it helps to resolve physical inconsistencies such as floating, colliding, or out-of-boundary objects.

\begin{table}[h]
    \centering
    \caption{
    \textbf{Ablation study.} We present in the following table the effect of removing different components from our method—Planner (\textit{Plan}), Hierarchical Placement (\textit{H-Place}), and Hierarchical Graph (\textit{H-Graph})—as well as our joint scene optimization (\textit{Opt.}). Our approach, which combines localized planning and hierarchical placement with joint scene optimization, achieves the best balance in geometric plausibility. 
    }
    \label{tab:ablation}
    \resizebox{\linewidth}{!}{\begin{tabular}{lcccccccc}
        \toprule
          Row ID& Opt. & H-Place & Plan & H-Graph & NoFloat (\%) $\uparrow$ & InBound (\%) $\uparrow$ & ColVol ($m^3$) $\downarrow$ & PIoU  $\downarrow$   \\
        \midrule
        
         1& \ding{55} & \ding{55} & \ding{55} & \ding{55} &52.02 & 62.93 & 1.3424 & 0.474 \\
         2& \ding{55} & \ding{55} & \ding{55} & \ding{51} &51.47 & 61.78 & 1.3473 & 0.475 \\
        
         3& \ding{55} & \ding{55} & \ding{51} & \ding{51} & 52.39 & 53.76 & \textbf{1.3301} &\textbf{0.465} \\
         4& \ding{55} & \ding{51} & \ding{55} & \ding{51} & 57.07 & 63.87 & 1.3302 &0.467 \\ 
         5& \ding{55} & \ding{51} & \ding{51} & \ding{51} & 77.08 & 90.08 & 1.3693 & 0.482 \\
         6& \ding{51} & \ding{51} & \ding{51} & \ding{51} & \textbf{88.41} & \textbf{99.77} & 1.3381 & 0.472  \\
        
        \bottomrule
    \end{tabular}}
\end{table}

\paragraph{Limitations.}
While our approach shows effective editing results for various complex real-world 3D scans, multiple limitations remain. 
In particular, while LLMs are quite powerful in determining which objects should be transformed, and potential coarse locations, they nonetheless still lack knowledge in more perceptual reasoning, which our 3D optimization constraints also do not consider. For instance, we can find possible locations for a vase on a table to move onto a shelf based on what would spatially fit, but we cannot account for which possibilities would be the most common sense ones or the most aesthetically pleasing. 

\begin{figure*}[tp]
    \centering
    \includegraphics[width=\linewidth]{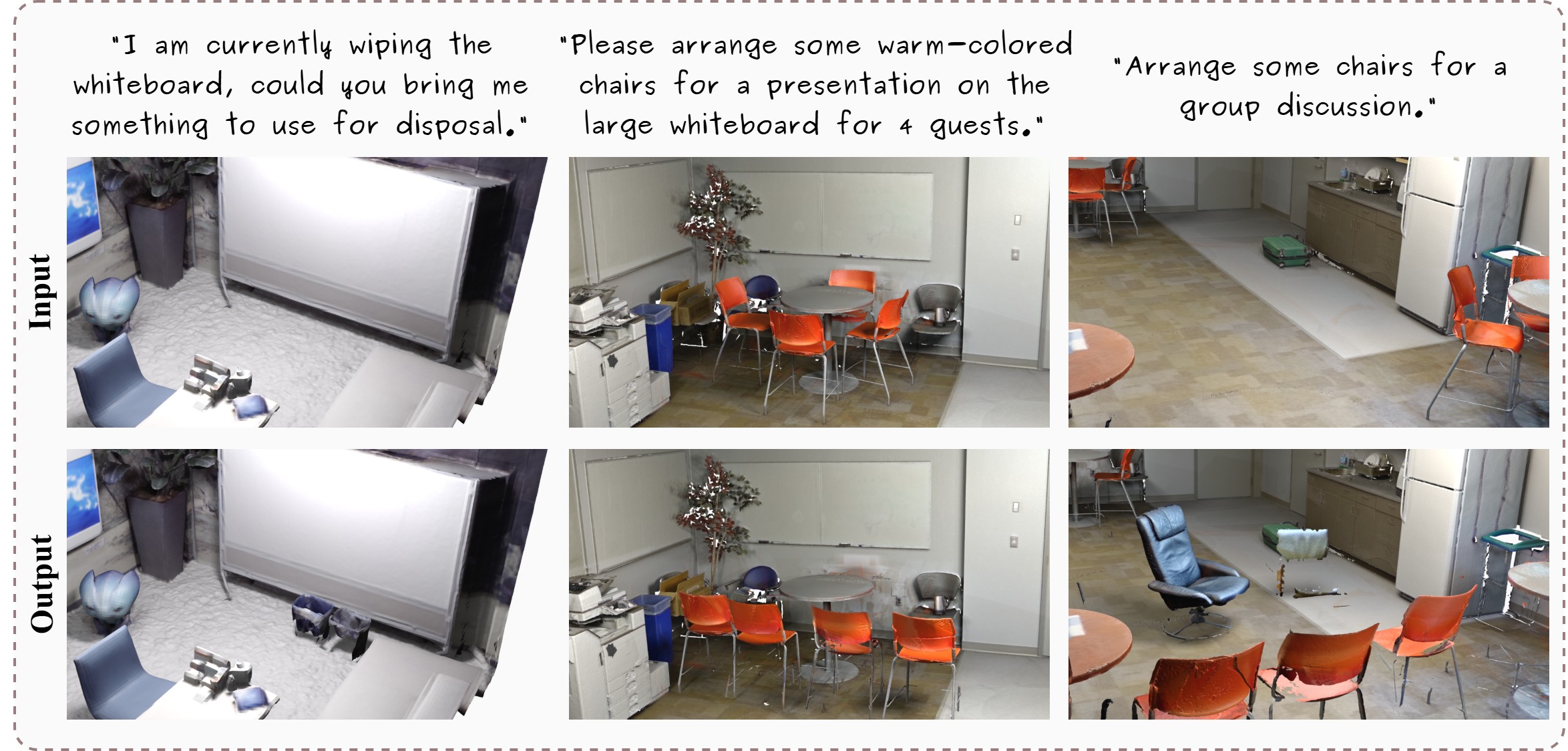}
    \caption{Editing results of our method with instance segmentation masks generated by Mask3D \cite{schult2023mask3d}, and trained on ScanNet++ \cite{yeshwanthliu2023scannetpp} training set. We show the results on two scenes, one from Replica \cite{straub2019replica} and ScanNet++ \cite{yeshwanthliu2023scannetpp} validation set.}
    \label{fig:ablation}
\end{figure*}
\section{Conclusion}
\label{sec:conclusion}
We have introduced \OURS{}, a hierarchically-guided approach to decompose high-level, natural language edits for 3D scenes into structured, holistic scene edits. 
This enables editing of complex, real-world 3D scans composed of hundreds of objects. 
Our use of LLM-generated high-level scene constraints together with 3D spatial constraints enables produced re-arrangements of 3D scans that achieve physically and semantically plausible edited 3D scans.
We believe this represents an important step towards natural editing of complex 3D scans for various content creation scenarios.

\section*{Acknowledgements}
This work was supported by the ERC Starting Grant SpatialSem (101076253).

\clearpage
{
    \small
    \bibliographystyle{ieeenat_fullname}
    \bibliography{main}
}

\clearpage
\setcounter{page}{1}
\maketitlesupplementary

\section{Additional Perceptual Study Details}
In our perceptual study, we conducted both a binary and a unary perceptual evaluation to assess the quality of edited 3D scenes. The unary perceptual study required participants to score each generated scene on two key criteria: \textbf{Adherence to Instruction} and \textbf{Layout Quality}. Participants were presented with a single edited 3D scene and asked to rate it on a scale from 1 (\textbf{Strongly Disagree}) to 5 (\textbf{Strongly Agree}) for each criterion. \textbf{Adherence to Instruction} evaluates how well the edited scene aligns with the given text instruction, ensuring that modifications accurately reflect the specified changes. \textbf{Layout Quality} assesses the overall spatial arrangement and positioning of objects, considering factors such as coherence, realism, and usability within the scene. This evaluation provides a fine-grained assessment of different methods' ability to generate high-quality and instruction-consistent scene modifications.

\section{Geometric Evaluation Metric Details}
\subsection{Collision metric (ColVol)}
\begin{figure}[h]
    \centering
    \includegraphics[width=\linewidth]{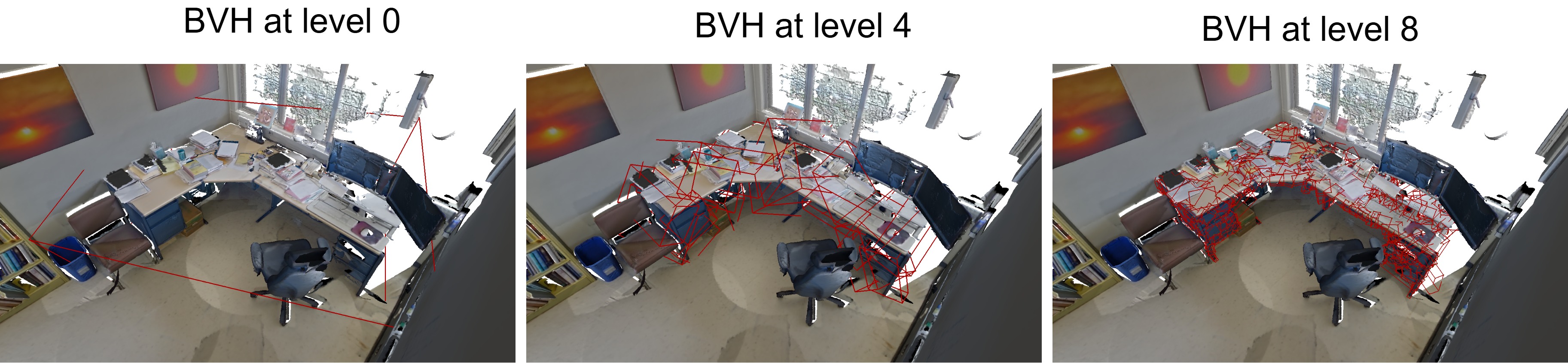}
    \caption{We show in this figure the bounding volumes of the instance object table at different levels in the bounding volume hierarchy BVH. We compute the colliding volumes between objects at level 8 as the sum of intersections in cube meters $m^3$.}
    \label{fig:bhv}
\end{figure}

First, we construct a bounding volume hierarchy with depth of 8 for each object in the scene to approximate its shape using bounding boxes at multiple levels (We show in Fig. \ref{fig:bhv} the visualization of bounding volumes hierarchy at different levels). 
We estimate the colliding volumes between pairs of objects using the sum of intersections of bounding volumes at level 8 in the bounding volume hierarchy. 

\subsection{In boundry metric (InBound)} 
For all objects that are moved in the edited version of the scene, we report the percentage of points which are inside the flour contour. For this, we compute the signed distance using the floor contour points and floor contour normals.

\subsection{Percentage of non floating objects (NoFloat)}
For each object that has been moved, we check if it is supported by a support surface or not with 1cm threshlod. An object is considered floating if it is elevated more than 1cm distance from the nearest support surface. In NoFloat metric we report the percentage of objects that are not floating.

\section{Estimating graph edges with 3D heuristics}

\subsection{Estimating `on top of' support surface relation}
We assign each object in the graph nodes \( N \) to the closest support surface, provided that the difference between the object's minimum height and the surface is less than 5 cm.

\subsection{Estimating `against wall'  relation}
An object $o_i$ is defined as against a wall $o_j$ if its front normal $\Vec{F}_i$ aligns with the front normal $\Vec{F}_j$ of the wall and the minimum distance between the wall and the object is less than 5cm.

\subsection{Estimating `facing' relation}
An object $o_i$ is defined to be 'facing' another object $o_j$ if the front normal $\Vec{F}_i$ of $o_i$ is aligned with the directional vector $\Vec{d} = c_{o_j}-c_{o_i}$, where $ c_{o_i}$  is the center of object $o_i$.

\section{Subgraph identification}
In the subgraph identification phase, we use an LLM, \( \Phi \), to reduce the set of objects to only relevant classes using prompt \ref{sec:class_ret}. Then, within these selected classes, we retrieve relevant nodes based on attributes like color, material, and description using prompt \ref{sec:inst_ret}. Next, we identify the key edges with prompt \ref{sec:edge_ret} that correspond to spatial instructions—for example, an instruction like "organize the top of the cabinet" refers to the 'on top of' edge, with the cabinet serving as its reference node.
\subsection{Class pruning prompt}
\label{sec:class_ret}



\begin{lstlisting}[
  basicstyle=\footnotesize\ttfamily, % Monospace font
    emph={list_of_class_names, Examples},           % Words to emphasize
  emphstyle=\bfseries\color{black},
  frame=single,
  breaklines=true,
  backgroundcolor=\color{white},
  showstringspaces=false
]
Relevant class selection prompt: 

You will be given a list of class names and your role is to locate the classes of interest. The classes of interest are objects to be moved and potential target locations. The classes don't have to be explicitly mentioned in the prompt; you must infer them from the context, especially the target locations where objects should be moved.

Among the following classes <objects>list_of_class_names</objects>, which ones are relevant to "{instruction}"?

Please give only the necessary class names to execute the task and make sure to return them in the tag:
<objects>[place list of objects here]</objects>.

Tips:
- Relevant classes do not necessarily need to be explicitly mentioned in the instruction. You should analyze the instruction and determine the most logical target location based on its context. For example, if the instruction is to "clear the inside of the fridge," and the class list includes a kitchen counter, it should also be considered. The kitchen counter is the most logical place to place the items, as it fits the context of the task.
- You must consider only class requirements; don't try to look for objects based on spatial relations like "close to" or "on top of." For example, "empty the inside of the fridge" means that the relevant classes are fridge and a potential location where to place the items, like a kitchen counter or table.
- Your role is to identify the most relevant classes based solely on semantics, without considering spatial relationships. For example, in the instruction "Move objects from inside the cabinet," your focus should be on the cabinet and a suitable target surface for placing the objects, such as a table or the floor. Do not attempt to determine which specific objects are inside the cabinet from the given list.

Important:
- The classes you return should be in the provided list of class objects; don't hallucinate any new class names.
- For every object class to be moved, you must return the most logical target class where it is going to be placed relative to either "close to" or "on top of" one of its support surfaces.



Examples of objects relevent to instruction:    
{Examples}

Very important: return classes from the provided list, don't hallucinte classes outside of it. If the instruction indicates prular objects you ust retrieve the closest from the set  <objects>{list_of_class_names}</objects>.

- The final output must be in XML format:

\end{lstlisting}

\subsection{Instance retrieval with node attributes}
\label{sec:inst_ret}

\begin{lstlisting}[
  basicstyle=\footnotesize\ttfamily, % Monospace font
    emph={Objects_details, Input_instruction, Examples},           % Words to emphasize
  emphstyle=\bfseries\color{black},
  frame=single,
  breaklines=true,
  backgroundcolor=\color{white},
  showstringspaces=false
]
You are a helpful assistant responsible for filtering out irrelevant objects based on some color, material, or size that might be mentioned in a given instruction. You will receive a list of objects, each with an ID, color, and material. Your task is to identify and return the object IDs that align with the instruction. If the instruction does not specify any attributes for the desired objects, return all object IDs.

Additionally, some objects serve as target locations for placing other objects according to the instruction. If no specific attributes are mentioned for these target objects, they should be retained.
If no color or material are requested in the instructions return all IDs.


Note: please select object that need to be moved as well as target objects.
Objects list with their object ids, class name, list of major colors, list of materials:

{Objects_details}

Instruction is :
{Input_instruction}

Please end your thinking with <relevant_ids>[place the list of ids which align with the instruction here]</relevant_ids>

Tips: 
    - Relevant classes do not necessarily need to be explicitly mentioned in the instruction. You should analyze the instruction and determine the most logical target location based on its context. For example, if the instruction is to "clear the inside of the fridge," and the class list includes a kitchen counter, it should also be considered. The kitchen counter is the most logical place to place the items, as it fits the context of the task.
    - You must consider only semantic attributes like color, material, and specific object characteristics as filtering requirements, don't try to look for objects based on spatial relations like close to or on top of. For example 'empty the inside of the fridge' means that the relevent classes are fridge and potential location where to place the items like kitchen counter or table.

Very Important:
    - Don't filter any object out unless a color or material attribute is requested.
    - Your task is to return the IDs of objects strictly from the provided list. For each object selected, you must include at least one target object ID from the list to indicate its placement relative to another object (e.g., near the door, on the table, or on the floor).

Examples of objects relevent to instruction:
    {Examples}
\end{lstlisting}

\subsection{Identifying relevent edges}
\label{sec:edge_ret}

\begin{lstlisting}[
  basicstyle=\footnotesize\ttfamily, % Monospace font
    emph={Objects_details, Input_instruction, Examples},           % Words to emphasize
  emphstyle=\bfseries\color{black},
  frame=single,
  breaklines=true,
  backgroundcolor=\color{white},
  showstringspaces=false
]
You are a Helpful Assistant. Your task is to determine how an object should be retrieved based on the given instructions. Do not try to make a plan for placing the objects. You must focus only on understanding whether retrieval involves relations or not.
    Guideline for Identifying the Object to Move and Retrieval Type

        Identify the Object to Move
            Look for action verbs like move, place, put to determine what is being acted upon.
            Example: "Move the chair." (Chair is the object to move).

        Check for Relations
            If a prepositional phrase (e.g., next to the table) describes the destination, it is retrieval without relations ("Move the chair next to the table.").
            If the phrase describes the current position, it is retrieval with relations ("Move the chair that is next to the table.").

        Determine Retrieval Type
            Without Relations: The object is retrieved by intrinsic properties (e.g., color, type).
            With Relations: The object is retrieved using another object as a reference.

        Edge Cases
            If no reference is mentioned, assume retrieval without relations.
            Multiple objects should be checked for independent or dependent relationships.

    Your role is to analyze retrieval type, not to decide where to place objects.

You will receive a list of objects, each with details such as their IDs, colors, and materials. The instruction provided may ask for objects relative to others in this list. Your task is to analyze the instruction and identify the objects that can serve as references to locate the target objects.

For example:
    If the instruction is: "Could you please empty the kitchen counter, then move items from the fridge to the kitchen counter?", the fridge has a relation "on top" with all possible surfaces, and the kitchen counter is also referenced as "on top." In this case, return the relevant object IDs based on these relationships.

Objects list with their object ids, class name, list of major colors, list of materials:

{Objects_details}

Instruction is :
{Input_instruction}

Please end your thinking with the following xm format 
```xml
<objects>
    <object>
        <id>[please the object id here]</id>
        <surface_ids>[please place the desired surface ids here seperated by a comma]</surface_ids>
        <relation>[please place the required relations here seperated by comma, you must choose from 'on top', 'facing', 'against wall']</relation>
    </object>            
</objects>

            ```

Important regarding surface IDs: A surface is the area where objects can be placed in an object, each object has surfaces with several IDs from 0 and up, the surface with the lowest elevation has the highest ID.
Example for surfaces: if there are three surfaces with ids 0,1,2,3 the lowest id corresponds to the top surface in this case ID 0 where the highest ID corresponded to the lowest surface in this case 3.

Examples that can help you understanding if the instruction requires retrieving with spatial constraints or not: 
    {Examples}
        
Hint how to address this task assigned to you: First, evaluate the instruction to determine whether any objects need to be retrieved in relation to others. If no objects are to be retrieved in relation to others, return an empty XML tag as follows: xml<objects></objects>.


\end{lstlisting}

\section{Prompts for planner}
In the planning phase, we generate first a plan where the LLM $\Psi$ generates a detailed plan while considering different target locations which define the hypotheses for moving each object, then it selects the best one while taking into account physical plausibility in support surfaces (max height does not accommodate the object to be moved or the surface is full). 
\subsection{Prompt for generating a plan}
\begin{lstlisting}[
  basicstyle=\footnotesize\ttfamily, % Monospace font
    emph={Objects_details, Input_instruction, Support_surfaces_details, Examples},           % Words to emphasize
  emphstyle=\bfseries\color{black},
  frame=single,
  breaklines=true,
  backgroundcolor=\color{white},
  showstringspaces=false
]
You will be given a list of objects with their ids colors and materials, you have to suggest a plan on how to place objects to execute the instruction.

Important analysis before planning how objects should be moved:
    Analyze the instruction and keep the movement of objects to the minimum, while insuring the instruction is satisfied. For example if the goal is to create a seating area to whatch TV, the TV should not be moved.

    
Objects list with their object ids, class name, list of major colors, list of materials:

{Objects_details}

Instruction is :
{Input_instruction}

Important: enphasis on objects that should not be moved in the plan, an example is " place the chairs to watch TV" the TV should remain untouched and the chairs should be placed in front of it while facing it.
Please end your thinking with <placement_plan>[place your detailed plan by specifying object ids and class names here]</placement_plan>
Important: enphasis on objects that should not be moved in the plan, an example is " place the chairs to watch TV" "arrange chairs for presentation on a screen" the TV and screen should remain untouched and the chairs should be placed in front of it while facing it.
{Support_surfaces_details}

Important when suggesting the plan: 
    - please suggest a target location for every object you want to move relative to other objects or the floor, even if the instruction is vague. 
        For example if an instruction says to clear a table, you must identify what objects are on the table and suggest new locations for these objects 
    - First analyze all potential target location, and place object relative to the most logical targets. Example, an instruction 'move the chair' has multiple targets but the most logical one is close to a table. 
    - If the target location is support object, please make sure to suggest which surface among the object surfaces it should go to in the plan
    - Some objects are better stacked on each other, for example papers and boxes shoulkd be stacked on each other if they are to be organized. where the largest one with dx,dy is the first.
    - The object level instruction must refer to an object (if desired to be placed close to it) or one of its surfaces (if desired to be placed on a surface) that exists in the list of objects.
    - Make the plan with natural language only, don't suggest to place objects in specific coordinates.
    - It is very important to pay attension to objects coordinates, since there are multiple objects with same functionality but different sizes. e.g. a large plant cannot be placed on top of a table, but a midsized or small plant one can be which gives a better vibe to the space.
    
When should you stack objects on top of each other:
    - Some objects like papers or boxes are better stacked on top of each other, in this case the plan should be:
        - place largest box on some surface 0 of another object or floor.
        - place second smallest box on the placed box.
        - and so on ... 
    - If three objects a (largest),b(smallest),c(medium) are to be stacked on top of each other a should be on a surface floor, or table, etc as it is the base of the stack, c should be on surface ID 0 of a, and b should be on surface ID 0 of c.   

When to use the coordinates:
    - Each objects has x,y coordinates use them to figure out far or close objects if mentioned in the instruction. It is strict to not include any coorcinates in the final placement plan.

How to handle instructions with few details:
    - Try to make a plan that can be physically plausible, for example placing 10 sofas for a presentation can not be done in one row, you have to place them in multiple rows in this case 3 rows would be good. 
\end{lstlisting}
\subsection{Prompt for converting the plan into hierarchical graph}
\begin{lstlisting}[
  basicstyle=\footnotesize\ttfamily, % Monospace font
    emph={placement_plan, dependency_plan, Examples},           % Words to emphasize
  emphstyle=\bfseries\color{black},
  frame=single,
  breaklines=true,
  backgroundcolor=\color{white},
  showstringspaces=false
]
You will be given a list of objects with their ids colors and materials and a plan to place these object. Your role is to return the logical dependency between objects and format the placement of objects in a hierarchical manner starting from the floor. Please strictly follow the placement plan

Important analysis before planning how objects should be moved:
    Analyze the instruction and keep the movement of objects to the minimum, while insuring the instruction is satisfied, pay attention to the placement plan, some objects are best to be kept unchouched you need to reflect that in the object level instruction in the hierarchy. For example if the goal is to create a seating area to whatch TV, the TV should not be moved. In this case the TV will nest the objects for seating, while having the instruction "keep the TV untouched".


Placement plan: 'place the table with id 10 near the door id 12 and a bottle with id 1 on top of the table (surface ID 0 of the table) with id 10, and the chair id 50 facing the table.'

Hierarchical Structure for Object Placement

    The hierarchy follows a nested dependency model, where objects are placed relative to their parent objects. This ensures spatial constraints are logically maintained.
    1. Root Level (Floor)

        The floor is the base of the environment, meaning all objects are ultimately placed on it.
        The floor itself remains static and untouched, serving as the foundational layer for all placements.

    2. First Nested Level (Door)

        The door (ID 12) is placed directly on the floor, meaning it is positioned independently.
        Since the door is static like the floor, it does not move or act as a container for other objects.

    3. Second Nested Level (Table)

        The table (ID 10) is placed near the door (ID 12).
        This means the table's position is spatially related to the door but not contained within it.
        Since the table is a movable object, its placement depends on the door's position.

    4. Third Nested Level (Chair & Surface)

        The chair (ID 50) is placed facing the table (ID 10), making it dependent on the table for its orientation.
        The table s surface (ID 0) is an implicit subcomponent of the table and serves as a placement area for smaller objects.
        While the surface is not a separate object, it acts as a reference point for placing items on the table.

    5. Fourth Nested Level (Bottle)

        The bottle (ID 1) is placed on top of the table (specifically, surface ID 0).
        Since the surface belongs to the table, the bottle is indirectly dependent on the table s placement.

    Purpose of the Hierarchy

    The structure enforces a logical dependency between objects. For example:
        The bottle s placement depends on the table.
        The table s placement depends on the door.
        The chair and surface placement depends on the table.
        The door s placement depends on the floor.

    This hierarchical representation reflects real-world relationships and ensures clarity in placement instructions.


How to approach this:
- You need to figure out the group center and what objects but be placed relative to the group centers.
- The group centers are placed relative to the floor, while the group memebers are placed relative to the group centers.
- If an object is placed in relation to other object it should be nested in it.


Final format (You must follow this output format):
xml ```
    <object>
        <name>floor</name>
        <id>floor_id</id>
        <instruction>leave the floor untouched as floors cannot be moved</instruction>
        <object>
            <name>door</name>
            <id>12</id>
            <instruction>leave the door untouched</instruction>
            <object>
                <name>table</name>
                <id>10</id>
                <instruction>place the table close to the door with id 12</instruction>
                <object>
                    <name>surface</name>
                    <id>0</id>
                    <instruction>Leave surface ID 0 untouched as it is part of the table</instruction>
                        <object>
                            <name>bottle</name>
                            <id>1</id>
                            <instruction>place the bottle on top of the surface with ID 0</instruction>
                        </object>
                </object>
                <object>
                    <name>chair</name>
                    <id>50</id>
                    <instruction>place the chair id 50 in front of the table, facing the table</instruction>
                
                </object>

            </object>
        </object>
    </object>
```

    
Extremely Important for the nesting:
    Even if an object doesn't move it must nest related objects.

Now please proceed with the following:


Placement plan :
{placement_plan}

Denependency plan:
{dependency_plan}

Structured xml hierarchy:
Please proceed with the step by step thinking then end it with xml output here
Important notes: 
    - If an object is placed in relation to other object (on top of, close to, near etc) it should be directly nested in it.
    - Please nest the desired surface in its correspoding object parent if exists in the placement plan.
    - An object is nested only if it should remain untouched or placed relative to the parent, but not moved from the parent.
                Example when not to nest: the cup should not be nested in the fridge in 'move the cup from the fridge' 
                Example when to nest: the cup should be nested in surface id 1 and surface id 1 should be nested in the fridge in 'place the cup in surface 1 in the fridge' 

Important regarding object level instructions:
    - Add the word 'untouched' in the instruction if the object is a surface or should not be moved. Surface instruction should always be 'leave surface untouched'

Very important: 
    - You must return the structure that enforces a logical dependency (use the dependancy plan in order to figure out which object is nested in which) between objects to figure out which object relates to which object before generating the xml nests
    - If an object is not explecitly mentioned to be moved in the placement plan you should leave it untouched, you your role is to structure the placement plan in a nested structure following the dependency plan


    Extremely important note: You must move only the objects that are mentioned to be moved in the instruction. For example 'move glass to the fridge' means the fridge must stay in place 
\end{lstlisting}
\section{Prompt for hierarchical object placement}
\begin{lstlisting}[
  basicstyle=\footnotesize\ttfamily, % Monospace font
    emph={Parent_object_name, Parent_object_id, parent_object_details, objects_to_be_placed, floor_details,Examples},           % Words to emphasize
  emphstyle=\bfseries\color{black},
  frame=single,
  breaklines=true,
  backgroundcolor=\color{white},
  showstringspaces=false
]
You will be given a reference object and a list of objects that you need to place relative to a reference object. Your role is to think step by step and seggest new locations, orientations, and constraints for the list of objects.
    Each object has to be placed following its instruction, you must suggest 3D coordinate for the base of the object, an orientation of the object, and a list of constraints with respect to the reference object.

    The representation of the reference object:
        - The reference object is represented with 
            1. Its base coordinate, which represents the 3D coordinate of the object with the minimum elevation (z) in meters and center in x and y.
            2. Its dimensions which represent the height(following the z axis), the width(following the x axis), the depth(following the y axis) in meters
            3. Its orientation, which refers to the orientation of the object around the z axis, in degrees.
            4. Its surfaces which can be used for placing objects, each surface has an ID where id 0 represents the surface with the highest elevation, the elevation is in meters.
            5. List of objects that are on top of the object

    The representation of the List of object to be placed relative to the reference object {Parent_object_name} id {Parent_object_id}:
            1. its base coordinate, which represents the 3D coordinate of the object with the minimum elevation (z) in meters and center in x=0 and y=0.
            2. Its orientation, which refers to the orientation of the object around the z axis, in degrees.

    The possible list of constraints with respect to the reference object {Parent_object_name} id {Parent_object_id} are: 
        - in_surface : this constraints concerns only instructions that require placing objects inside or on top of a reference object, if the instruction requires placing an object on top, that means the surface ID is 0 since it is the one with the highest elevation
        - facing: This constraint concerns objects which should be facing the reference object in a natural setting, for example a chair should be facing a referebce object whiteboaard.


    Important details when reporting the final location and orientation:
        - Don't perform the math operation instead report the formula with values and operations to get the new location or orientation. The allowed operations are :*: for multiplication, /: for division, -: for minus, +: for summation, cos: for cosine function,  sin: for sin function. angles should be in degrees.

    Hint on relations between objects:
        - If an object is facing another, its orientation should be 180 degrees minus the orientation of the other object
        - if an object is facing the same direction as another, both should have the same orientation

    Important details on the reference object's {Parent_object_name} id {Parent_object_id} orientation and location:
        - the reference object is oriented towards (meaning its front) the positive direction of the x-axis. And its base coordinate is located near the origin. 


    Output format:
        Please end your step by step thinking with the following xml which summarizes the new locations , new orientations and list of constraints of objects with respect to the reference object relative to the reference object {Parent_object_name} id {Parent_object_id}:
    xml ```
    <objects>
        <object>
            <id>[please place the object ID here]<id>
            <name>[please place the object name here]<name>
            <new_base_coordinate>[please place here the new base coordinate of the object]</new_base_coordinate>
            <new_orientation>[please place here the new orientation the object]</new_orientation>
            <constraints>
                <facing>[please mention here wheather the object should be facing the reference object {Parent_object_name} id {Parent_object_id} or not, answer with yes or no]<facing>
                <in_surface>[If the object should be in one of the surfaces place the surface ID here, if not place None]<in_surface>
                <distanace_to_reference>[please place the distance to the reference object here]<distanace_to_reference>
            </constraints>
        <object>
    <objects>

    ```  

    Reference Object: 
    {parent_object_details}

    List of Objects to be placed with their instructions: 
    {objects_to_be_placed} 

    {floor_details}

    Now please proceed with your suggested new coordinates and orientations, taking into account that the reference object x center and y center are both 0,0 and this is similar to its base coordinate. It is also important to note that all objects are represented with base coordinate which is the center of the object in x and y and minimum elevation in z.

    Important: the front of the reference object  is in the positive x axis, thus if an object should be in front of it it must be placed at x>0 and y that is bounded.
    Important: you must use place the objects elevation (z axis) at the floor level if the object should be on the floor.
    Important regarding in_surface constraint: if the reference object is a surface of an object, please put the ID of the surface (which is a number, don't put something like fridge's surface or table's surface) in this constraint.

    Very important: Pay attention to the instruction for each object, if an object should not be moved please don't include it in the xml list. If no object should be moved return an empty xml <objects></objects>


    Left, right, front, and back sides conventions:
        Left side with respect to the reference object is : y < 0
        Right side with respect to the reference object is : y > 0
        Front side with respect to the reference object is : x > 0
        Back side with respect to the reference object is : x < 0

    Very important when placing objects against walls or other objects:
        If an object is placed against a wall or another object <facing> should be no and the orientation of the object to be place should be the same as the reference object 
        If an object is against the wall, its x should be half the dx dimension of the object and y can range from -dy_wall/2 to dy_wall/2

    Very important for the <facing> constraint which are with respect to the reference object {Parent_object_name} id {Parent_object_id}: 
        If an object should be facing the same direction as the reference object {Parent_object_name} id {Parent_object_id} <facing> constraint should be set to no 
        Example (helpful for reasoning only):
            - place chairs with respect to another reference object chair for a role that requires them to be in the same direction like guest watching TV, means that <facing> is no
            - place chairs with respect to another reference object TV for a role that requires them to be in the same direction like guest watching TV, means that <facing> is yes 
\end{lstlisting}


\end{document}